\documentclass[journal]{IEEEtran}

\usepackage[utf8]{inputenc}
\usepackage{amsmath}
\usepackage{times}
\usepackage{graphicx}
\usepackage{color}
\usepackage{multirow}
\usepackage[ruled]{algorithm}
\usepackage[noend]{algpseudocode}
\usepackage{pifont}
\usepackage{tabu}
\usepackage{array,multirow, makecell}
\usepackage{subcaption}
\usepackage{latexsym}
\usepackage{color,soul}
\usepackage{mathtools}
\usepackage{wrapfig}

\DeclarePairedDelimiter\floor{\lfloor}{\rfloor}

\begin{document}

%

\title{Discretization based Solutions for Secure Machine Learning against Adversarial Attacks
}
%
%
%
%
%

%
\author{Priyadarshini~Panda, \textit{ Student Member}, \textit{IEEE}, Indranil~Chakraborty,
        and Kaushik~Roy,~\IEEEmembership{Fellow,~IEEE}
\thanks{P. Panda, I. Chakraborty and K. Roy are with the School
of Electrical and Computer Engineering, Purdue University, West Lafayette,
IN, 47907 USA E-mail: (pandap, ichakra, kaushik)@purdue.edu.}
}

\maketitle

\begin{abstract}
Adversarial examples are perturbed inputs that are designed (from a deep learning network’s (DLN) parameter gradients) to mislead the DLN during test time. Intuitively, constraining the dimensionality of inputs or parameters of a network reduces the ‘space’ in which adversarial examples exist. Guided by this intuition, we demonstrate that discretization greatly improves the robustness of DLNs against adversarial attacks. Specifically, discretizing the input space (or allowed pixel levels from 256 values or $8\-bit$ to 4 values or $2\-bit$) extensively improves the adversarial robustness of DLNs for a substantial range of perturbations for minimal loss in test accuracy. Furthermore, we find that Binary Neural Networks (BNNs) and related variants are intrinsically more robust than their full precision counterparts in adversarial scenarios. Combining input discretization with BNNs furthers the robustness even waiving the need for adversarial training for certain magnitude of perturbation values. We evaluate the effect of discretization on MNIST, CIFAR10, CIFAR100 and Imagenet datasets. Across all datasets, we observe maximal adversarial resistance with $2\-bit$ input discretization that incurs an adversarial accuracy loss of just $\sim1-2\%$ as compared to clean test accuracy. 
\end{abstract}

%
%
%

%
%

%
%




\section{Introduction}
Deep Learning Networks (DLNs) have exhibited better than human performance in several vision-related tasks \cite{lecun2015deep}. However, they have been recently shown to be vulnerable toward adversarial attacks \cite{kurakin2016adversarial, goodfellowexplaining, panda2018explainable}: slight changes of input pixel intensities that fool a DLN to misclassify an input with high confidence (Fig. 1). What is more worrying is that such small changes (that craft adversaries) are visually imperceptible to humans, yet, mislead a DLN. This vulnerability severely limits the potential safe-use and deployment of DLNs in real-world scenarios. For instance, an attacker can fool a DLN deployed on a self-driving car to mispredict a STOP sign as a GO signal, and cause fatal accidents.

Subsequently, there have been several theories pertaining to the adversarial susceptibility of DLNs \cite{goodfellowexplaining}. The most common one suggests that the presence of adversary is an outcome of the excessive linearity of a DLN (a property of high-dimensional dot-products). While one can argue that ReLU-type activation imposes non-linearity in a model, the linear operations such as Convolution, Pooling exceed the number of non-linear ReLU operations. Further, ReLU is typically a linear functionality in the $>0$ regime, and hence, plagues a DLN to be sufficiently linear. Now, this linearity causes a model to extrapolate its behavior for points in the hyper-space (of data and model parameters) that lie outside the training/test data manifold. Adversarial inputs, essentially, are images that are synthesized such that they ‘lie far’ from the typical data manifold and hence get misclassified. 
\setlength{\textfloatsep}{6pt plus 1.0pt minus 8.0pt}
\begin{figure}
\centering
\includegraphics[width=0.7\linewidth]{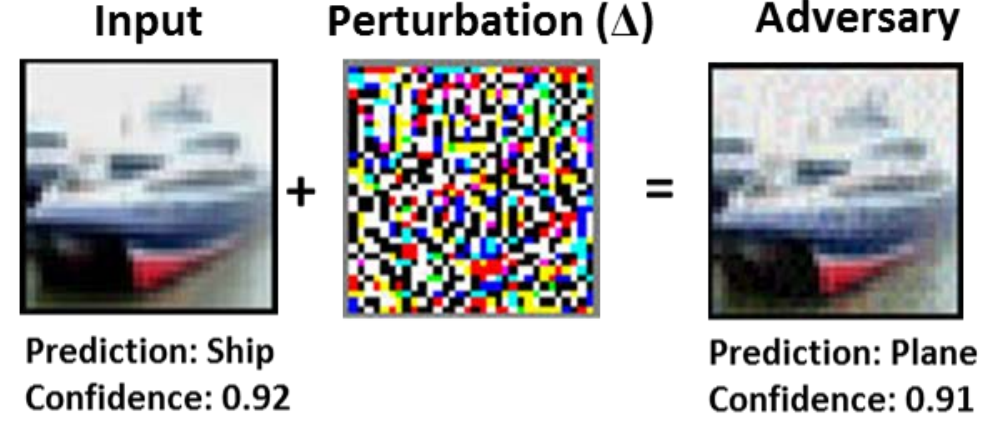}
\caption{\textit{An image of a \textbf{ship} perturbed with adversarial noise yields an adversarial image that fools the classifier. The classifier predicts the original image correctly with a confidence of 92\%, while gets fooled by the adversarial image mispredicting it as \textbf{plane} with a high confidence of 91\%.}}
\end{figure}

Fig. 2 (a) demonstrates this data manifold intuition and adversarial input creation with a cartoon. Since DLNs are discriminative models, they partition a very high-dimensional input space into different classes by learning appropriate decision boundaries. The class-specific decision boundaries simply divide the space into hyper-volumes. Interestingly, these hyper-volumes encompass the training data examples as well as large areas of unpopulated space that is arbitrary and untrained. This extrapolation of decision boundary beyond the training data space is a result of \textit{‘linearity’}, that in turn, gives rise to generalization ability. The fact that a model trained only on training data is able to predict well on unseen test data (termed as, generalization) is a favorable outcome of this extrapolation property. Unfortunately, this property also exposes a model to adversarial attacks. Adversarial data are created by simply adding small perturbations to an input data point, that shifts it from its manifold (or hyper-volume) to a different hyper-volume (that the model has not been trained upon and shows extrapolated behavior), causing misclassification. 

From the above intuition, one can deduce that adding regularization features to a DLN’s training will improve its generalization ability and in turn, adversarial robustness. In fact, the most effective form of adversarial defense so far is training a model with adversarial data augmentation (called adversarial training) \cite{szegedy2013intriguing}. It is evident that explicit training on adversarial data will increase the model’s capability to generalize and hence predict correctly on unseen adversarial data. However, the above discussion on excessive linearity and hyper-space dimensionality points to an alternate and unexplored regularization possibility, that is discretizing or constraining the data manifold for achieving adversarial robustness. For instance, discretizing the input data (say from 256-pixel value levels (or 8-bit) to 4 levels (or 2-bit)) reduces the regions into which data can be perturbed. In other words, the minimum perturbation required to shift a particular data point from one hyper-volume to another will increase in a discretized space (Fig. 2 (b)). This in turn will intrinsically improve the resistance of a DLN. Similarly, discretizing the parameter space (as in binarized neural networks (BNNs) \cite{hubara2016binarized}) will introduce discontinuities and quantization in the manifold (that is non-linear by nature). This will further decrease the extent of hyper-volume space that is arbitrary/untrained and thus reduce adversarial susceptibility (Fig. 2(b)). It is evident that such discretization methods have an added advantage of computational efficiency. In fact, low-precision neural networks (BNNs and related XNOR-Nets \cite{rastegari2016xnor}) were introduced with a key motif of obtaining reduced memory and power-consumption for hardware deployment of DLNs. 
\setlength{\textfloatsep}{5pt plus 1.0pt minus 4.0pt}
\begin{figure}
\centering
\includegraphics[width=\linewidth]{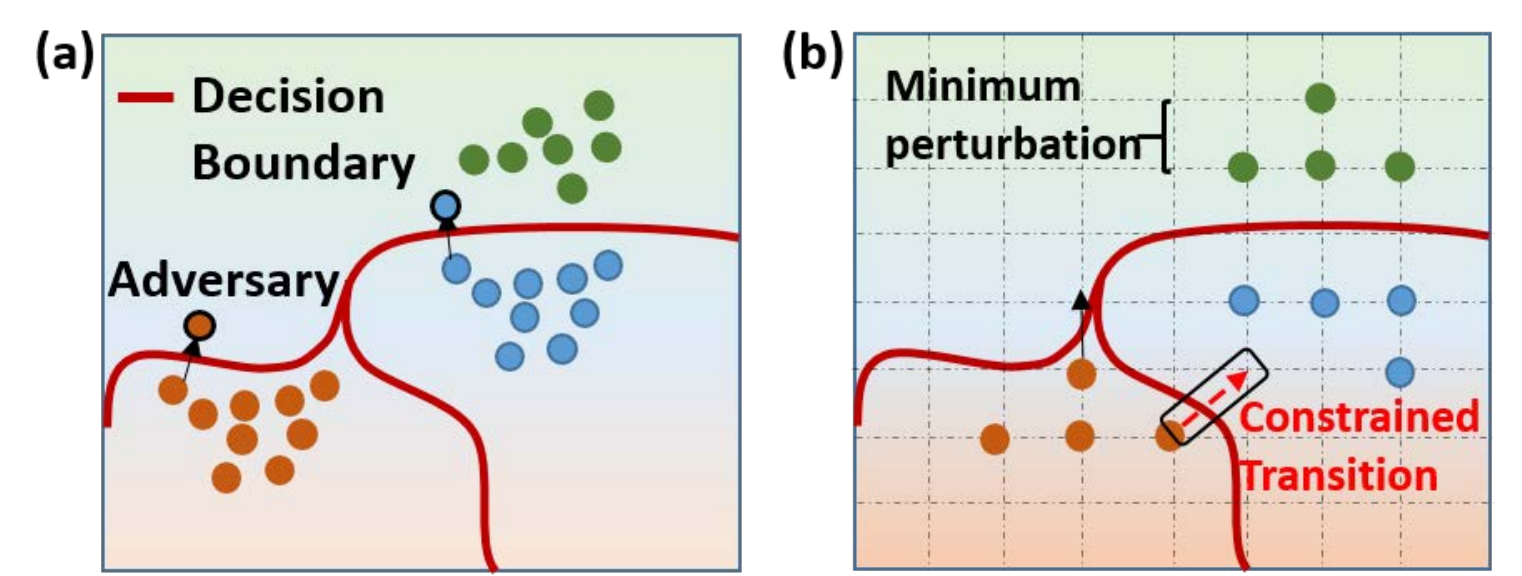}
\caption{\textit{Cartoon of the intuition behind adversary creation and discretization. (a) The data points (shown as 'dots') encompass the data manifold in the high-dimensional subspace. The classifier is trained to separate the data into different categories or hyper-volumes based on which the decision boundary is formed. Note, the decision boundary is a characteristic of the trained parameters (weights) of the model. The decision boundary is, however, extrapolated to vast regions of the high-dimensional subspace that are unpopulated and untrained because of \textbf{linear} model behavior. Adversaries are created by perturbing the data points into these empty regions or hyper-volumes and are thus misclassified (\textbf{orange} mispredicted as \textbf{green} in this case). (b) Discretization quantizes the data manifold thereby introducing a minimum perturbation required to shift a data point. As quantization will increase, so will the minimum allowed distortion. Further, discretization constrains the creation of adversaries since not all transitions can cause a data point to shift between hyper-volumes.}}
\end{figure}

In this paper, we demonstrate that discretization, besides offering obvious efficiency improvements, has far-reaching implication on a model’s adversarial resistance. We particularly emphasize on three different discretization themes and illustrate their suitability toward improving a DLN’s adversarial robustness, as follows:

\begin{itemize}
\item{\textit{Discretization of input space:} We reduce the input dimensionality by quantizing the RGB pixel intensities into a variable range: $2^8 =256$ to $2^2=4$. We show that for minimal loss in accuracy, the adversarial robustness of a model substantially improves ($<1\%$ accuracy difference between clean test and adversarial test data), even, without any adversarial training. Furthermore, we show that combining adversarial training with 2-bit input discretization makes a model substantially more robust (than adversarial training with full 8-bit input precision) for large perturbation ranges.}
\item{\textit{Discretization of parameter space:} We show that models trained with low-precision weights and activations, such as BNNs, are intrinsically more robust to adversarial perturbations than full precision networks. Furthermore, we find that training BNNs with adversarial data augmentation is difficult. However, increasing the capacity of the BNN (with more neurons and weights) minimizes the adversarial training difficulty. For sufficient model capacity, adversarially trained BNNs yield higher adversarial robustness than their full-precision counterpart.}
\item{\textit{Discretization of both input \& parameter space:} We demonstrate that combining input discretization with binarized weight /activation training greatly improves a model’s robustness. In fact, training a BNN with input discretization (say, 2-bit input) yields similar or better adversarial accuracy as that of an adversarially trained full-precision model. Thus, the combined discretization scheme can be seen as an efficient alternative to achieving adversarial robustness without the expensive data augmentation procedure.}
\end{itemize}

\section{Related Work}
Based on the intuition demonstrated in Fig. 2, robustness of DLNs can be attributed to two factors: property of the input and model property. Consequently, there have been many recent proposals \cite{guo2018countering, buckman2018thermometer, chen2018improving,  xu2017feature} that exploit the input-dependent factor and try to remove adversarial perturbations by applying input preprocessing or transformations. Due to the simplicity of this approach, these methods are attractive for practical implementations as they do not incur large computational overhead (as with adversarial training) and do not interfere with the learning process. Our paper complements the results of the prior works while presenting a novel result on the effectiveness of combined parameter and input discretization for adversarial robustness.

One of the first works on input discretization by Xu et al. \cite{xu2017feature} propose a \textit{depth-color-squeezing} technique wherein they reduce the degrees of freedom available to an adversary by `removing' unnecessary features. Our pixel discretization scheme is based on their color depth reduction technique. However, the key idea in \cite{xu2017feature} is to compare the model's prediction on the original input with its prediction on the squeezed input during testing. Xu et al. train the model with regular inputs and during inference use pixel discretization to detect adversarial inputs. That is, if the original and squeezed inputs produce predictions with large difference (greater than an user-defined threshold), the model deems the input to be adversarial and rejects it. Ultimately, the model outputs prediction for only legitimate or non-adversarial inputs. In contrast, the key novel aspect of our work is to train a model with discretized pixel data such that the model looks at a reduced input subspace during training that decreases or constrains its' ability to overly genralize in the high-dimensional subspace. Similarly, the thermometer encoding technique and input transformation technique proposed in \cite{buckman2018thermometer}, \cite{guo2018countering} are guided by the same intuition of reducing adversarial subspace dimensions. Guo et al. \cite{guo2018countering} trained the network with images transformed in various ways and observed improved adversarial resistance. However, they measured robustness for controlled graybox attack settings (where, the model parameters are known to the attacker but the input transformations are unknown). Our paper's results on whitebox attack is a stronger notion of robustness as we assume all parameters as well as input discretization known to the attacker. Thus, our results while supporting the claims of \cite{guo2018countering} are more substantial and generalizable. In Buckman et al. \cite{buckman2018thermometer}, the authors propose a thermometer encoding technique to map input pixels to a binary vector in order to make more meaningful change during pixel discretization without losing any information from the original image. While, the authors show good adversarial robustness results on small tasks, such as, MNIST \cite{lecun1998mnist}, they are shown to achieve poor performance on more complex datasets (like, CIFAR10 \cite{krizhevsky2009cifar}) \cite{chen2018improving}. 

In order to address the limitation of the prior works solely based on input space modifications, we investigate the effect of combining model discretization with the input transformation property thereby leveraging both criteria that contribute to adversarial dimensions. To the best of our knowledge, we are the first to formally evaluate and analyze the impact of input and parameter space discretization for DLNs (across simple and complex datasets including ImageNet2012 dataset) on robustness. A recent work \cite{galloway2018attacking} demonstrated the effectiveness of BNNs against adversarial attacks and observed a similar difficulty in adversarial training with BNNs. However, they did not consider input space discretization and its impact on robustness. While complementing their results, we show that quantizing the input pixels of a BNN during training greatly improves its robustness, even waiving the need for the expensive and time-consuming adversarial training, for certain perturbation ranges. 

\section{Background on Adversarial \\Attacks} 
\textit{Generating Adversaries :} Adversarial examples are created using a trained DLN’s parameters and gradients. As shown in Fig. 1, the adversarial perturbation, $\Delta$, is not just some random noise, but carefully designed to bias the network’s prediction on a given input towards a wrong class. Goodfellow et. al \cite{goodfellowexplaining} proposed a simple method called Fast Gradient Sign Method (FGSM) to craft adversarial examples by linearizing a trained model’s loss function ($\mathcal L$, say cross-entropy) with respect to the input ($X$):
\begin{equation}
X_{adv} = X + \epsilon \times sign(\nabla_X \mathcal L(\theta, X ,y_{true}))
\end{equation}
Here, $y_{true}$ is the true class label for the input $X$, $\theta$ denotes the model parameters (weights, biases etc.) and $\epsilon$ quantifies the magnitude of distortion. The net perturbation added to the input ($\Delta = \epsilon \times sign(\nabla_X \mathcal L(\theta, X ,y_{true}))$) is, thus, regulated by $\epsilon$. Distorting the input image in the direction of steepest gradient has the maximal effect on the loss function during prediction. Intuitively, referring to Fig.2, this distortion shifts the data point from the trained region or hyper-volume to an arbitrary region thereby fooling the model. 

\textit{Types of Attacks :} In machine learning literature, there are two kinds of attacks: Black-Box (BB), White-Box (WB) that are used to study adversarial robustness \cite{tramer2018ensemble}. WB adversaries are created using the \textit{target} model’s parameters, that is, the attacker has full knowledge of a target model’s training information.  BB attacks refer to the case when the attacker has no knowledge about the target model’s parameters. In this case, adversaries are created using a different \textit {source} model’s parameters trained on the same classification task as the target model. Since BB attacks are transferred onto the target model, they are weaker than WB attacks. Security against WB attacks is a stronger notion and robustness against WB attacks guarantees robustness against BB for similar perturbation ($\epsilon$) range. 

\textit{Adversarial Training :} Adversarial training simply injects adversarial examples into the training dataset of a model \cite{szegedy2013intriguing}. For each training sample in the dataset, an adversary is created using FGSM \cite{goodfellowexplaining}. There are several forms of adversarial training. For instance, instead of using the same $\epsilon$ for all training examples, \cite{kurakin2016adversariala, tramer2018ensemble} propose to sample a unique $\epsilon$ (from a random normal distribution) for each training example. This will increase the variation in the adversaries created thereby increasing the robustness of a network to larger range of $\epsilon$ values. The authors in \cite{madry2018towards} use WB adversaries created, using a multi-step variant of FGSM to guarantee a strong defense against both BB and WB attacks. Note, the common theme across all adversarial training methods is data augmentation.

In this work, we focus on adversarial attacks created using FGSM and evaluate the robustness of models against WB adversaries. We evaluate a model’s robustness/report adversarial accuracy on the adversarial dataset created using the test data for a given task. 

\section{Experiments} 
We conduct a series of experiments for each discretization theme, primarily using MNIST \cite{lecun1998mnist} (Fully Connected Network, FCN) and CIFAR10 \cite{krizhevsky2009cifar} (AlexNet \cite{gitbnn} architecture), detailing the advantages and limitation of each approach. We compare the adversarial robustness of each discretization approach with its’ full-precision counterpart (with and without adversarial training), using $\epsilon$ values reported in recent works \cite{tramer2018ensemble,madry2018towards}. For adversarial training, we employ Random-step FGSM (R-FGSM) proposed in \cite{tramer2018ensemble} to create a variety of training set adversaries. R-FGSM perturbs the input $X$ with a small random step (sampled from a normal distribution $\mathcal N$) before adding the loss gradient to the input: $X_{adv} = X’ + \Delta$, where $X’ = X + \alpha sign(\mathcal N(0, I), \alpha = \epsilon/2$. We use WB adversarial training to confer strong robustness toward all forms of attacks. Note, for MNIST (CIFAR), we use $\epsilon = 0.3$ $(8/255)$ during adversarial training. For evaluating the robustness of parameter space discretization, we use BNNs \cite{hubara2016binarized, gitbnn} to evaluate CIFAR10 and MNIST datasets. We also evaluate the robustness of discretization methods on large-scale datasets, CIFAR100 (ResNet20 architecture \cite{he2016deep}) and Imagenet \cite{deng2009imagenet} (AlexNet architecture) using XNOR networks \cite{rastegari2016xnor, gitxnor}. Please note, for MNIST we use two different FCN architectures: FCN1-4 hidden layer network with 6144 neurons each (784-6144($\times$4)-10), FCN2-4 hidden layer network with 600 neurons each (784-600($\times$4)-10). We imported github models \cite{gitxnor, gitbnn} for implementing our experiments. We used the same hyperparameters (such as weight decay value, learning rate etc.) as used in \cite{gitxnor, gitbnn} to train our models. It is worth mentioning that our paper is the only other work besides \cite{galloway2018attacking} demonstrating the effectiveness of discretized/binarized parameter space on adversarial attacks. While \cite{galloway2018attacking} conducted experiments with various forms of attacks (primarily, on MNIST), we restrict ourselves to the WB attack scenario and extrapolate our analysis on larger datasets.

\subsection{Discretization of Input Space} 
\setlength{\textfloatsep}{5pt plus 1.0pt minus 4.0pt}
\begin{figure}
\centering
\includegraphics[width=0.85\linewidth]{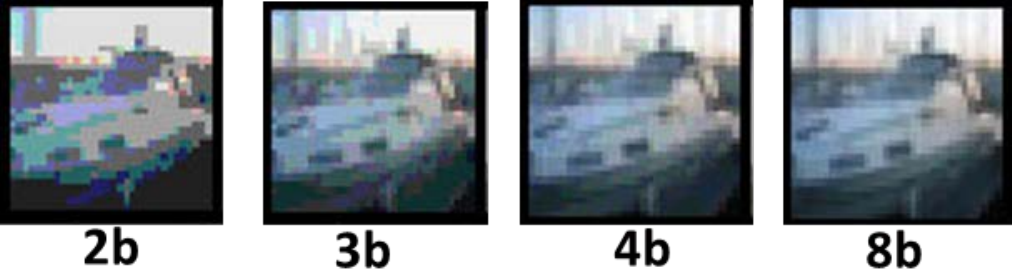}
\caption{\textit{Sample images from CIFAR10 dataset for varying levels of input pixel discretization: \textbf{$2-bit, 3-bit, 4-bit, 8-bit$}}}
\end{figure}
With input space discretization, we convert the raw integer pixel intensities ($0 \le I \le 255$) that are typically 8-bit (or $8b$) values to a low precision range of $\delta$ bits ($0 \le I_{\delta} \le 2^\delta$) as:
\begin{equation}
I_\delta = \floor*{\frac{I}{(256/2^\delta)}} (\frac{256}{2^\delta}) + \frac{1}{2} (\frac{256}{2^\delta})
\end{equation}
where $\floor*{}$ denotes integer division. Such quantization reduces the number of data points (given a grayscale input image of size $N\times N$) in the manifold from ${{2^8}^{N^2}}$ to ${{2^\delta}^{N^2}}$. This can be broadly interpreted as reducing the redundancy in the input data. Fig. 3 illustrates sample CIFAR10 images discretized to varying $\delta$ values. The corresponding accuracy (trained on AlexNet for 20 epochs) is shown in Table \ref{table1}. There is a natural tradeoff between input discretization and overall accuracy of a network. Yet, the test accuracy loss from the full precision $8b$ to $3b$ is $\sim2.2\%$. This verifies the presence of unnecessary and redundant input dimensions that do not substantially contribute to the classification task or accuracy. $2b$ discretization decreases the accuracy by a larger margin ($\sim6\%$). Note, this accuracy loss can be minimized by training the $2b$ inputs for more epochs. However, for iso-comparison, we fix the number of epochs across all experiments for a given dataset. 
\setlength{\intextsep}{3pt}%
\setlength{\columnsep}{2pt}%
\begin{wraptable}{r}{4.5cm}
\setlength{\tabcolsep}{4pt} 
\caption{\textbf{CIFAR10 Accuracy}}
\label{table1}
\centering
\begin{tabular}{|c|c|p{\textwidth}}
\hline
Input-bit & Accuracy \\
\hline
2b & 82\\
\hline
3b & 86.64\\
\hline
4b & 87.1\\
\hline
8b & 88.9\\
\hline
\end{tabular}
\end{wraptable}
\setlength{\textfloatsep}{5pt plus 1.0pt minus 4.0pt}
\begin{figure}
\centering
\includegraphics[width=0.7\linewidth]{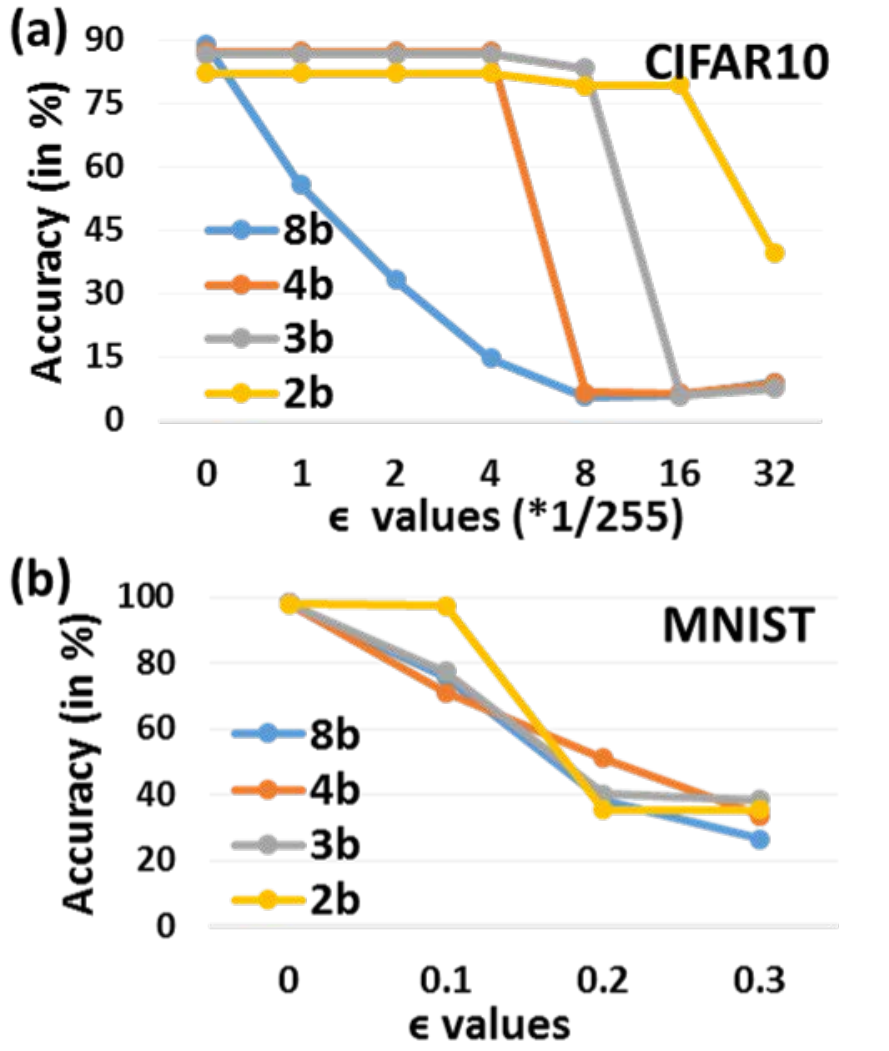}
\caption{\textit{Adversarial accuracy on test data for varying perturbation values on (a) CIFAR10 (b) MNIST for different input discretization}}
\end{figure}

A remarkable outcome of this discretization method is the substantial improvement in a model’s adversarial accuracy. Fig. 4 illustrates the evolution of adversarial accuracy of the CIFAR10 models (from Table \ref{table1}) with increasing level of perturbation, $\epsilon$. Note, $\epsilon = 0$ corresponds to clean test accuracy. It is clear that clamping the input dimensionality to lower values increases the resistance of the model to larger magnitude of distortion. We can thus deduce that removing the redundant dimensions in the input reduces the overall hyper-volume space thereby leaving ‘less’ space for shifting or adversarially perturbing a data point (referring to Fig.2 intuition). $8b$ input model shows a decline in accuracy even for a small value of $\epsilon =1/255$. This further confirms that presence of redundant input dimensions allow even small perturbations to shift a data point. In contrast, increasing discretization increases the minimum $\epsilon$ that affects a model’s accuracy catastrophically. What is surprising is that for $2b$ input, a model’s adversarial accuracy ($\sim79\%$) for large $\epsilon = (8,16)/255$ is almost similar to that of clean accuracy ($\sim82\%$). For larger $\epsilon = 32/255$, the accuracy of all models declines to $<10\%$, except $2b$. This is a very interesting result since we have not employed any adversarial training, and still achieve substantial adversarial resistance for a large $\epsilon$ range.

Fig. 4(b) shows the adversarial accuracy results for MNIST (trained on FCN2 for 10 epochs). We observe a similar trend of increasing adversarial resistance with increasing discretization for larger $\epsilon$. Since MNIST is a simple dataset with predominantly black-background, input discretization ($8b,4b,3b$) does not contribute much to adversarial resistance until we go to extremely low $2b$ precision. In fact, $2b$ discretization yields adversarial accuracy similar to the clean test accuracy (for $\epsilon=0.1$) exhibiting the effectiveness of this technique even for simple datasets.

Next, we trained the $2b$ input discretized CIFAR10 and MNIST models with adversarial training to observe the improvement in adversarial accuracy compared to $8b$ input adversarial training (Table \ref{table2}). Comparing to the results in Fig. 4(a, b), adversarial training substantially improves the robustness of a model with full $8b$ input for larger $\epsilon$ values. Input discretization greatly furthers this robustness with $>10\%$ accuracy gain across different perturbation ranges in both MNIST and CIFAR10. It is worth mentioning that the CIFAR10 accuracy (79\%) without adversarial training for $\epsilon =(8,16)/255$ for $2b$ input is as good as the accuracy (83\%) with adversarial training. This shows that input discretization is a good regularization scheme that improves the generalization capability of a network on adversarial data. Note, for $\epsilon =32/255$ in case of CIFAR10, the accuracy is similar for $8b, 2b$ since the adversarial training was conducted with adversaries created using $\epsilon =8/255$. Including larger perturbation adversaries during adversarial training will yield improved accuracy gain. 
\begin{table}
\caption{\textbf{Accuracy with adversarial training for varying $\epsilon$. Text in \textcolor{red}{red} are the $\epsilon$ values for MNIST and corresponding accuracy.}}
\label{table2}
\centering
\begin{tabular}{|c|c|c|c|c|c|p{\textwidth}}
\hline
Data & Model & Clean & \makecell{$\epsilon$:8/255\\ \textcolor{red}{0.1}} &  \makecell{$\epsilon$:16/255\\ \textcolor{red}{0.2}}&  \makecell{$\epsilon$:32/255\\ \textcolor{red}{0.3}} \\
\hline
\multirow{2}{4em}{CIFAR10}& 2b & 83.1 & 82.7 & 82.7 & 43.9 \\
& 8b & 84.3 & 62.2 & 53.6 & 45.5\\
\hline
\multirow{2}{4em}{MNIST}& 2b & 98.5 & \textcolor{red}{98.5} & \textcolor{red}{84.7} & \textcolor{red}{85.4} \\
& 8b & 98 & \textcolor{red}{84.8} & \textcolor{red}{74.5} & \textcolor{red}{65.9} \\
\hline
\end{tabular}
\end{table}

\subsection{Discretization of Parameter Space} 
Since input discretization gave us such promising results, we were naturally inclined toward analyzing a binarized neural network’s (BNN) behavior against adversarial attacks. Here, the weights and activations (or parameters) are discretized to extremely low precision values $\{+1, -1\}$ \cite{hubara2016binarized}. The discretization constraints are imposed on a BNN during training, wherein, the parameters are clamped to $\{+1, -1\}$ after every backpropagation step. One can view this discretization as an implicit form of regularization. In fact, it is this extreme form of regularization that makes a BNN difficult to train (clean test accuracy observed with BNNs is, typically, lower than full-precision networks). As suggested in \cite{galloway2018attacking}, the difficulty in training a BNN translates to difficulty in attacking the BNN as well. Referring to the data manifold intuition (Fig. 2), we can deduce that constraining the parameter space during a model’s training will introduce discontinuities and non-smoothness in its decision boundary. Since adversaries are created using gradients of a model (that is a property of the model’s decision boundary), generating gradients (and hence adversaries) for non-smooth functions will be difficult. This in turn will make a BNN less susceptible to adversaries. Note, the input image to a BNN is full $8b$ precision. 
\setlength{\textfloatsep}{3pt plus 1.0pt minus 4.0pt}
\begin{figure}
\centering
\includegraphics[width=0.8\linewidth]{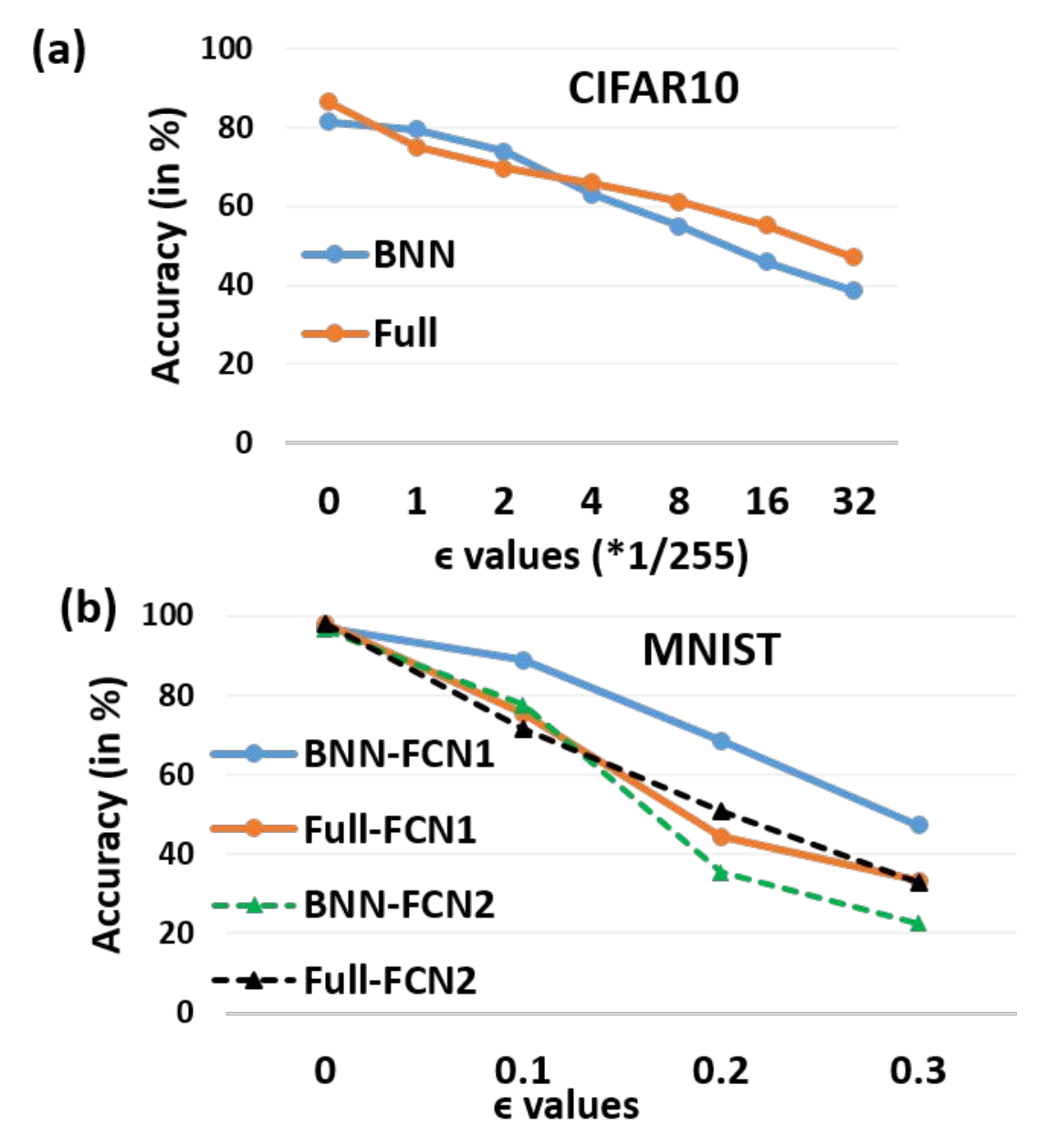}
\caption{\textit{Adversarial accuracy on test data for varying perturbation values on (a) CIFAR10 (b) MNIST for binarized and full-precision ($32b$ weights) models}}
\end{figure}

Fig.5(a) compares the adversarial accuracy obtained for varying $\epsilon$ values for CIFAR10 BNN  (AlexNet architecture) against a similar architecture full-precision network (with $32b$ precision for weights and activations). We trained the networks for 40 epochs since BNNs require more training iterations to attain comparable accuracy as that of a full-precision network. Here, we do not incorporate input discretization in our analyses. All networks are fed $8b$ inputs. In Fig. 5(a), for $\epsilon \le 2/255$, BNN shows better adversarial resistance (i.e. adversarial accuracy is closer to clean accuracy). However, the BNN’s accuracy declines steeply as we move toward larger perturbation ranges. We note a similar trend for MNIST (trained for 10 epochs on FCN2 architecture), wherein the full-precision network yields improved robustness than the BNN for $\epsilon \ge 0.2$. These results contradict our intuition that increased discretization of BNNs should result in lesser adversarial susceptibility. 

To understand this, we calculated the L1 norm of the first hidden layer activation of the FCN2 network in response to clean input images. We found that BNNs generally have a larger variance and range of values than full-precision network. Since BNN uses weight values (+/-1) which are typically of greater magnitude than the small weight values of a full precision network, we observe a larger range in the former case. Interestingly, we find that the L1 norm of the BNN activations (in response to adversarial images perturbed with lower $\epsilon$ values) approximately lie within the same range as that of the clean input case. In contrast, L1 norm for higher $\epsilon$ adversaries have a much higher range. For a full-precision network, the L1 norm range of the different $\epsilon$ adversaries and clean data typically intersect with each other owing to the lower weight values (Fig. 6). We believe that the extreme quantization of weight values in BNNs to higher magnitudes causes adversarial susceptibility for larger range perturbations. While the L1 norm analysis is not very substantial from a mathematical standpoint, it hinted us to increase the capacity (more neurons and weights) of the network. The motif here is that increasing the capacity would increase the overall range of activation values that might incorporate larger range perturbations. Exploding the network capacity for MNIST (FCN1 architecture) yielded a sizable improvement in adversarial resistance with BNN as compared to its corresponding full-precision counterpart (Fig. 5(b)). This is a crucial detail of our analysis that: \textit{while BNNs are intrinsically robust to adversaries (for small $\epsilon$), only models with sufficient capacity can withstand against large $\epsilon$ values}. 

\begin{figure}
\centering
\includegraphics[width=0.8\linewidth]{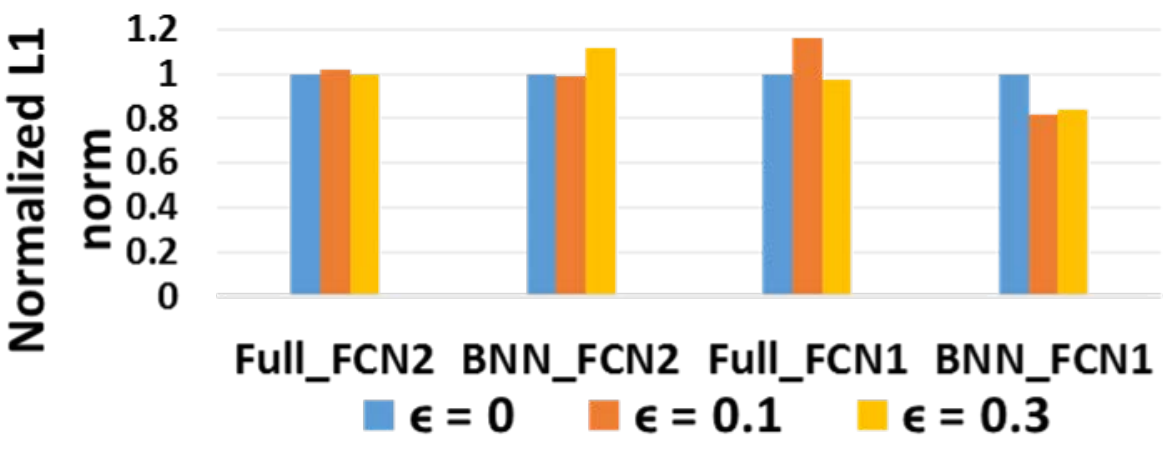}
\caption{\textit{Normalized L1 norm of first hidden layer activations in response to clean ($\epsilon=0$) and adversarial inputs ($\epsilon=0.1, 0.3$) for binarized and full-precision MNIST model of different architecture: FCN1, FCN2}}
\end{figure}

\setlength{\textfloatsep}{3pt plus 0.0pt minus 0.0pt}
\begin{table}
\caption{\textbf{Accuracy with adversarial training for varying $\epsilon$. Text in \textcolor{red}{red} are the $\epsilon$ values for MNIST and corresponding accuracy.}}
\label{table3}
\centering
\begin{tabular}{|c|c|c|c|c|c|p{\textwidth}}
\hline
Data & Model & Clean & \makecell{$\epsilon$:8/255\\ \textcolor{red}{0.1}} &  \makecell{$\epsilon$:16/255\\ \textcolor{red}{0.2}}&  \makecell{$\epsilon$:32/255\\ \textcolor{red}{0.3}} \\
\hline
\multirow{2}{4em}{CIFAR10}& BNN & 79.7 & 53.1 & 43.6 & 35.3 \\
& Full & 82.7 & 72.2 & 63.6 & 55.5 \\
\hline
\multirow{2}{4em}{MNIST (FCN1)}& BNN & 96.9 & \textcolor{red}{89.1} & \textcolor{red}{74.5} & \textcolor{red}{65.8} \\
& Full & 98 & \textcolor{red}{84.8} & \textcolor{red}{71} & \textcolor{red}{61.7} \\
\hline
\end{tabular}
\end{table}

Even with adversarial training, we observed the same trend that binarized networks of insufficient capacity do not yield as good adversarial robustness as that of a full-precision network (Table \ref{table3}). For CIFAR10, full-precision network is the clear winner. While for MNIST (with excessive capacity FCN1 architecture), BNN yields improved robustness. A noteworthy observation here is that adversarial training substantially improves the robustness of a full-precision network (see CIFAR10 results in Fig. 5(a) , Table \ref{table3}), while BNNs do not benefit much from them. In fact, we find that BNNs are difficult to train with adversarial training. The learning rate/other hyperparameters need to be tuned carefully to ensure that the BNN model converges to lower error values during adversarial training. \cite{galloway2018attacking} also observed a similar trend and explained that binarized weights are not as \textit{‘malleable’} as full-precision weights and hence cannot easily adjust to all possible variations of adversarial data augmented to the training dataset. We think that increasing the capacity of the network compensates for the \textit{`non-malleability'} of the constrained parameters to certain extent. As a result, we see improved accuracy for MNIST in Table \ref{table3} with FCN1 architecture.

\subsection{Discretization of Input and Parameter \\Space} 
Next, we combined both discretization strategies and analyzed the adversarial robustness of BNNs with varying image-level discretization. We compare the adversarial accuracy of BNNs  to that of a full-precision network for iso-input discretization scenarios, as shown in Table \ref{table4} for CIFAR10 (AlexNet architecture trained for 40 epochs). In Table \ref{table4}, \textit{BNN-2b} (\textit{Full-2b}) refers to a binarized (full-precision) model with $2b$ input precision. Full precision models have $32b$ precision weights and activations. While input discretization for a full-precision network suffers a sizeable accuracy loss, BNN’s accuracy fluctuation is marginal with a maximum of $1\%$ change. This is expected since BNNs (owing to +/-1 binarized parameters) do not have as many dimensions (as a full-precision network with $32b$ weights and activations) to fit the extra information in the $8b$ input data. Thus, BNNs fit $2b$, $8b$ data likewise yielding similar generalization error. As opposed to the results seen earlier with $8b$ inputs, BNNs with lower input precision ($2b, 4b$) have significantly higher adversarial resistance than their full-precision counterparts even for large $\epsilon$ values. \textit{Model capacity does not restrict the adversarial resistance in this case. This is an artefact of the two-step quantization that increases the minimum allowable perturbation to shift a data point.} We can also draw an alternate insight from this result: The constrained parameter space of BNNs restricts their overall exploration of the data manifold during training. Referring to Fig. 2 (b), this increases the probability of untrained or arbitrary hyper-volumes (for BNNs) thereby increasing their adversarial susceptibility. Increasing the capacity enables a BNN to explore the manifold better during training. By discretizing the input, we are restricting the overall data manifold space. This allows a model, even, with lower capacity to explore the manifold well thereby decreasing the extent of arbitrary hyper-volumes. Table \ref{table5} illustrates the accuracy results for MNIST (FCN1 architecture trained for 20 epochs).

\begin{table}
\caption{\textbf{Adversarial accuracy with CIFAR10 for varying $\epsilon$ with different combinations of input and parameter discretization.}}
\label{table4}
\centering
\begin{tabular}{|c|c|c|c|c|p{\textwidth}}
\hline
Model & Clean & \makecell{$\epsilon$:8/255} &  \makecell{$\epsilon$:16/255}&  \makecell{$\epsilon$:32/255} \\
\hline
BNN-2b & 81 & 80 & 80 & 36.7 \\
Full-2b & 82 & 79.1 & 79.3 & 39.6 \\
\hline
BNN-4b & 81.9 & 58.3 & 52.3 & 36.8 \\
Full-4b & 81.1 & 53.8 & 45.1 & 37.3 \\
\hline
BNN-8b & 81.5 & 55.1 & 45.9 & 38.6 \\
Full-8b & 86.5 & 61.1 & 55.2 & 47.1 \\
\hline
\end{tabular}
\end{table}

\begin{table}
\caption{\textbf{Adversarial accuracy with MNIST for varying $\epsilon$ with different combinations of input and parameter discretization.}}
\label{table5}
\centering
\begin{tabular}{|c|c|c|c|c|p{\textwidth}}
\hline
Model & Clean & \makecell{$\epsilon$:0.1} &  \makecell{$\epsilon$:0.2}&  \makecell{$\epsilon$:0.3} \\
\hline
BNN-2b & 96.4 & 96.4 & 60.7 & 62.3 \\
Full-2b & 97.8 & 97.4 & 35.4 & 35.3 \\
\hline
BNN-4b & 96.4 & 88.9 & 76.7 & 58.7 \\
Full-4b & 98.1 & 71.1 & 50.9 & 33.7 \\
\hline
BNN-8b & 97.1 & 89.4 & 56.1 & 33.6 \\
Full-8b & 98.2 & 75.9 & 38.5 & 26.4 \\
\hline
\end{tabular}
\end{table}

We conducted adversarial training with $2b$ input discretized BNNs to find out if it helps build adversarial robustness. The results are shown in Table \ref{table6}. Comparing to the $8b$ input BNN adversarial training results in Table \ref{table3}, we observe a substantial gain in adversarial accuracy. However, contrasting the BNN results against Table \ref{table2} ($2b$ input full-precision networks), we observe similar performance gains. In fact, the accuracy gains for $2b$ input CIFAR10 BNN with and without adversarial training (Table\ref{table4}/Table \ref{table6}) are nearly the same. Earlier, we saw that the accuracy (for low $\epsilon$ values) of a full-precision network working on $2b$ input data without adversarial training is similar to that of an adversarially trained network on $8b$ inputs (Table \ref{table2}, Fig. 4). Combining the adversarial training results till now, we can deduce the following: \textit{1) For low input-precision ($2b$) regime, adversarial training does not compound the adversarial resistance of a network (irrespective of binarized or full-precision parameters), for lower $\epsilon$ values.  Adversarial training helps when the input has higher ($8b$) precision. 2) Input discretization, in general, offers very strong adversarial defense  for lower $\epsilon$ values. Discretizing the input as well as the parameter space furthers adversarial robustness. Adversarial training in a discretized input and parameter space does not benefit much and hence can be waived.} However, in case of stronger multi-step attack scenarios and to gain robustness against larger perturbations (such as $\epsilon=32/255, 0.3$ in CIFAR10, MNIST), the network needs to be adversarial trained with corresponding large $\epsilon$ values. 
\begin{table}
\caption{\textbf{Accuracy with adversarial training for varying $\epsilon$. Text in \textcolor{red}{red} are the $\epsilon$ values for MNIST and corresponding accuracy.}}
\label{table6}
\centering
\begin{tabular}{|c|c|c|c|c|c|p{\textwidth}}
\hline
Data & Model & Clean & \makecell{$\epsilon$:8/255\\ \textcolor{red}{0.1}} &  \makecell{$\epsilon$:16/255\\ \textcolor{red}{0.2}}&  \makecell{$\epsilon$:32/255\\ \textcolor{red}{0.3}} \\
\hline
CIFAR10& BNN-2b & 78.4 & 78.1 & 78.1 & 30.5 \\
\hline
\makecell{MNIST\\(FCN1)}& BNN-2b & 95.7 & \textcolor{red}{95.7} & \textcolor{red}{89.3} & \textcolor{red}{88.6} \\
\hline
\end{tabular}
\end{table}
\begin{figure}
\centering
\includegraphics[width=0.75\linewidth]{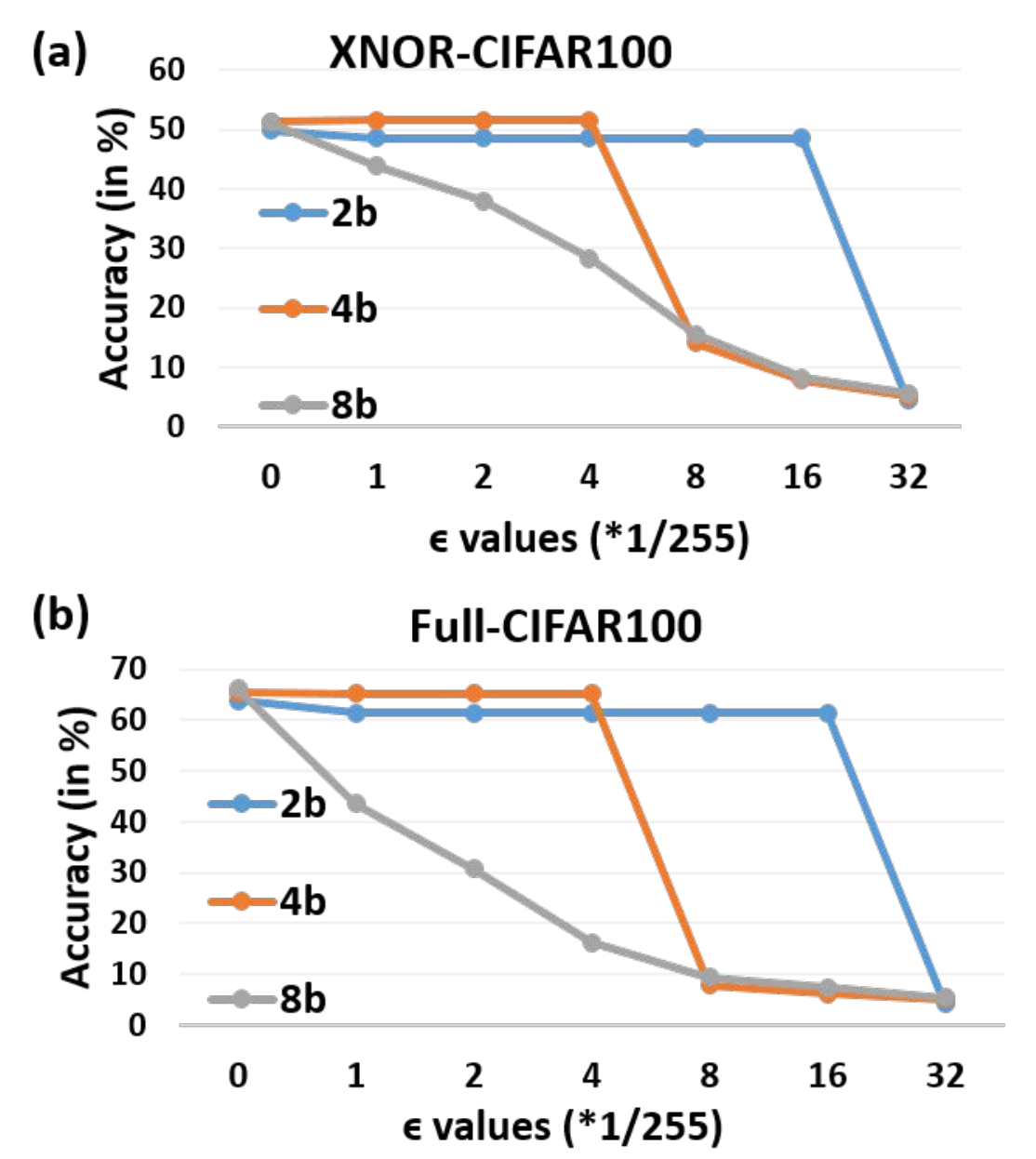}
\caption{\textit{Adversarial accuracy on test data for increasing $\epsilon$ values on binarized and full-precision ($32b$ weights) models trained on CIFAR100 with different input discretization:\textbf{$8b, 4b, 2b$}}}
\end{figure}
\subsection{Analysis on CIFAR100 and Imagenet}
Scaling up the discretization analysis to larger datasets yielded similar results as observed with CIFAR10, MNIST. Fig. 7 demonstrates the adversarial accuracy evolution for CIFAR100 (trained on ResNet20 architecture for 164 epochs) for binarized XNOR ($1b$ weights and activations) networks and corresponding full-precision ($32b$ weights and activations) models. Note, XNOR networks are similar to BNNs (1-bit weights/activations) with additional scaling factors to achieve higher accuracy on complex datasets. It is evident that input discretization is the most beneficial to obtain adversarial robustness. $2b$ input discretized models in both cases yield adversarial accuracy close to the clean accuracy ($\epsilon =0$) for a large range of perturbations. The accuracy loss between clean and $\epsilon=16/255$ adversary for $2b$-input XNOR ($1.6\%$) is slightly better than the $2b$-input full-precision model ($2.4\%$). This can be attributed to the intrinsic robustness offered by discretizing the parameter space of XNOR networks. Furthermore, the fact that $8b$-input XNOR yields higher adversarial accuracy for iso-perturbation values than $8b$-input full-precision model further demonstrates the ability of binarized networks to counter adversarial attacks. A noteworthy observation here is that the loss in clean accuracy between $2b$ and $8b$ input discretized full-precision network is small ($\sim2\%$) as compared to the large $6\%$ loss observed earlier with CIFAR10 (Table \ref{table1}). As we scale up the complexity of the dataset, the redundancies in the input dimensions increase. Discretizing the input for a complex dataset thus eliminates more redundant dimensions that do not contribute to the accuracy. In contrast, smaller datasets have lesser redundant dimensions and are thus at a risk of suffering a large accuracy drop with input discretization. 
\setlength{\textfloatsep}{5pt plus 1.0pt minus 4.0pt}
\begin{figure}
\centering
\includegraphics[width=0.75\linewidth]{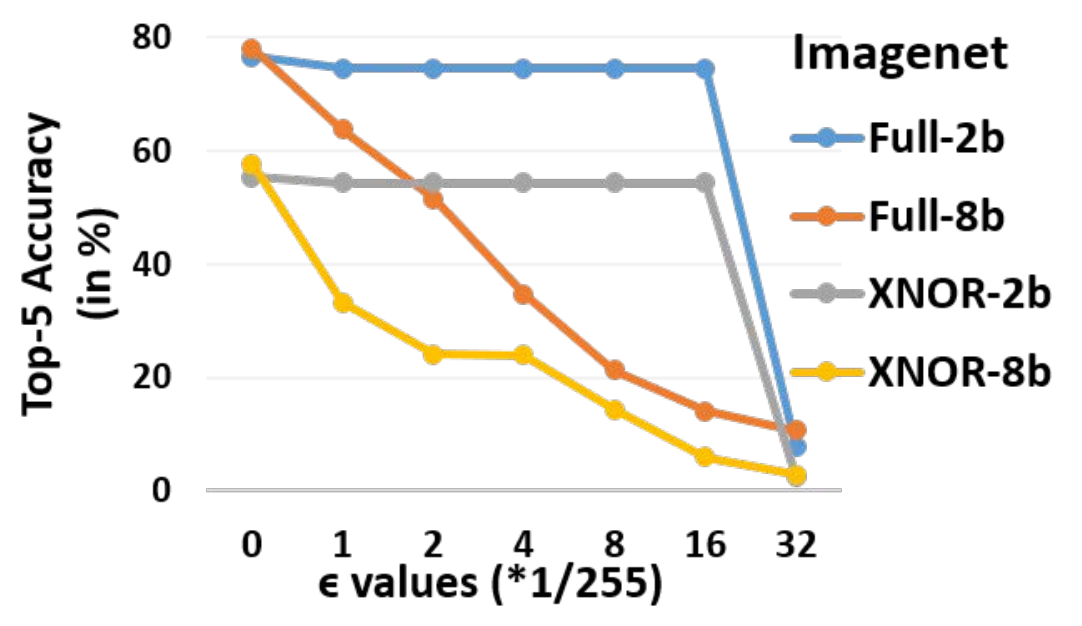}
\caption{\textit{Top-5 Adversarial accuracy on test data for increasing $\epsilon$ values on binarized and full-precision ($32b$ weights) models trained on Imagenet with different input discretization:\textbf{$8b, 2b$}}}
\end{figure}

Fig. 8 shows the accuracy results for Imagenet (trained on AlexNet). We only show the top-5 adversarial accuracy. We see similar trends as CIFAR100. Note, the XNOR models are trained for 50 epochs, while full-precision models are trained for 90 epochs. As a result, we see lower baseline accuracy ($\epsilon=0$) in the former case. Like CIFAR100, the loss in clean test accuracy between $2b$ and $8b$ input discretization is minimal for each model. Also, the accuracy difference between clean and $\epsilon=16/255$ adversarial data for $2b-XNOR$ ($0.9\%$) is much lower than $2b-full$ precision models ($2.8\%$). This highlights the intrinsic robustness capability of binarized networks even for large-scale datasets.  

\section{Conclusion}
Low-precision models or quantization techniques, so far, have been explored to reduce the resource utilization of DLNs for energy-efficient deployment on edge devices. We have demonstrated that \textit{discretization} also warrants \textit{security} against adversarial attacks, thereby, offering a key benefit of \textit{robustness} in hardware implementation. In summary, the main findings/recommendations from this work are:
\begin{itemize}
\item{Input discretization is major benefactor for adversarial robustness (with both binarized and full-precision) models. $2b$ input discretized models (without adversarial training) yield similar adversarial accuracy as adversarially trained $8b$ input models for lower $\epsilon$ values. Robustness against higher $\epsilon$ and multi-step attack requires adversarial training.}
\item{Binarized (low-precision weights/activations) models are intrinsically more (although, slight) robust than full-precision ($32b$ weights/activations) models. Adversarial training needs to be carefully done on sufficient capacity binarized networks to attain similar adversarial robustness as the full-precision models.} 
\item{Combining input and parameter discretization is an efficient way of obtaining adversarial robustness to a moderate range of perturbation values without conducting the iterative adversarial training.}
\end{itemize}
Our work unravels a simple idea that: \textit{hardware optimization related techniques can potentially resolve or resist software vulnerabilities (specifically, adversarial attacks)}. While we focus on discretization, there is a lot of future scope to explore other efficiency-driven techniques (such as, stochasticity, model pruning etc.) to gauge their implication on adversarial robustness.

\section*{Acknowledgment}
This work was supported in part by C-BRIC, Center for Brain-inspired Computing, a JUMP center sponsored by DARPA and SRC, by the Semiconductor Research Corporation, the National Science Foundation, Intel Corporation and by the Vannevar Bush Faculty Fellowship.

\end{document}